# Computing With Words for Student Strategy Evaluation in an Examination

Prashant K Gupta        Pranab K. Muhuri

Department of Computer Science, South Asian University,
New Delhi-110021, India

**Abstract-** In the framework of Granular Computing (GC), Interval type 2 Fuzzy Sets (IT2 FSs) play a prominent role by facilitating a better representation of uncertain linguistic information. Perceptual Computing (Per-C), a well-known computing with words (CWW) approach, and in its various applications have nicely exploited this advantage. This paper reports a novel Per-C based approach for student strategy evaluation. Examinations are generally oriented to test the subject knowledge of students. The number of questions they are able to solve accurately judges success rates of students in the examinations. However, we feel that not only the solutions of questions, but also the strategy adopted for finding those solutions are equally important. More marks should be awarded to a student, who solves a question with a better strategy compared to a student whose strategy is relatively not that good. Furthermore, the student's strategy can be taken as a measure of his/ her learning outcome as perceived by a faculty member. This can help to identify students whose learning outcomes are not good and thus, can be provided with any relevant help, for improvement. The main contribution of this paper is to illustrate the use of CWW for student strategy evaluation and present a comparison of the recommendations generated by different CWW approaches. CWW provides us with two major advantages. Firstly, it generates a numeric score for the overall evaluation of strategy adopted by a student in the examination. This enables comparison and ranking of the students based on their performances. Secondly, a linguistic evaluation describing the student strategy is also obtained from the system. Both these numeric score and linguistic recommendation are together used to assess the quality of a student's strategy. Furthermore, the linguistic recommendation is useful for human beings as they naturally understand and express themselves using 'words', 'words' being treated as fuzzy information granules in the GC paradigm, which is perhaps the case with most of the human reasoning and concepts. Also, through the comparison of the recommendations generated by different CWW approaches, we found that Per-C outperforms the others CWW approaches by generating unique recommendations in all the cases as well as modeling the word uncertainty in the best possible way.

## 1 Introduction[1]

Lin (1997) first coined the term 'Granular Computing (GC)'. Since then GC has become quite popular and been used in a number of applications (Yao 2005; Yao 2016). GC (Zadeh 1997; Zadeh 1998) deals with representation and processing of information granules (Pedrycz and Chen 2015a). Information granules (Liu et al. 2016) are the collection of individual pieces of information, arranged together by virtue of similarity, indistinguishability, etc. (Pedrycz 2001). Information granules occur frequently in everyday life (Pedrycz and Chen 2015b). For example, a person's name may be granulated as first name, middle name and surname. These information granules may be precise when they represent numeric data, say height of a person.

However, a large amount of information in real life situations is in the linguistic form. Linguistic information contains 'words' and sentences drawn from natural language (Zadeh 1975). For example, consider a person, who goes to a supermarket to buy a dress. Before making a decision about whether to buy or not a dress, he/ she evaluates it on various parameters like fit, cost, etc. According to the person, the fit of the dress can be 'very comfortable', 'comfortable', etc. and the cost can be 'very high', 'affordable', etc. After evaluation, the person may describe the dress as 'good', 'very good', 'average', etc. The words used above like 'very comfortable', 'comfortable', 'very high' are all examples of linguistic information or 'words' or linguistic terms. Collection of such linguistic terms is a linguistic term set.

As computers are increasingly pervading all known walks of life, using them to process the linguistic information inevitably calls for the design of systems which perform complex tasks like human beings. However, human beings understand and express themselves naturally in terms of 'words'. Therefore, to design a system that works in a manner similar to the human beings, it is desirable that the computing system must process the linguistic information like human beings[2].

---

Email addresses: guptaprashant1986@gmail.com (Prashant K Gupta),  pranabmuhuri@cs.sau.ac.in (Pranab K. Muhuri).

[2] Human beings process linguistic information seamlessly due to the capability of human cognitive process.

Linguistic information has a remarkable characteristic of being imprecise and vague (Zadeh 1975) because it depends on a user's perception, which is subjective. This can be summed up by the adage 'words mean different things to different people' (Mendel and Wu 2010). Fuzzy sets (FSs) (Zadeh 1965), can be used to capture and model the 'word' uncertainty in the best possible way (Zadeh 1998; Gupta et al. 1979; Zadeh 1997). There are different types of FSs viz., type-1 FSs or type-2 FSs, etc. and the words represented using the FSs are the uncertain information granules (Pedrycz and Chen 2011).

There are number of application areas where the concepts FSs have been applied. Castillo et al. (2016a) used a combined approach of GC and fuzzy logic to solve the problem of aircraft control. They used the GC to divide the problem framework into granules, and fuzzy logic to the control problem. Castillo et al. (2016b) compared the performance of fuzzy logic controllers based on different types of FSs using four benchmarks. In (Sanchez 2015), authors presented a methodology for construction of IT2 fuzzy information granules using the uncertainty-based information. IT2 FSs were used to represent component reliability in the multi-objective reliability-redundancy allocation problem for improved trade-off solutions (Muhuri et al. 2017). Cervantes and Castillo (2015) presented a novel architecture for design of fuzzy controllers. They designed the bigger fuzzy controller as a combination of simple individual fuzzy controllers, and then used fuzzy aggregator to obtain the combined output. In (Sanchez et al. 2015), authors developed a generalized type 2 fuzzy logic controller for robotic application and compared its performance against type 1 and interval type 2 fuzzy logic controllers. Castillo and Melin (2008) said that to handle information uncertainty in the real life problems in the best possible way, use of IT2 FSs is inevitable. They also said that T1 FSs are insufficient to handle such uncertainties.

Chen et al. (2001) presented a novel approach for generating fuzzy rules from numerical data, for fuzzy classification problems. Chen et al. (2009) presented a methodology to forecast the student enrolment in a university using the clustering and fuzzy logic relationships. Chen and Chen (2011) presented a new method for three applications viz., forecast the Taiwan Stock Exchange Capitalization Weighted Stock Index (TAIEX), the enrollments of the University of Alabama and the inventory demand based on high-order fuzzy logical relationships. The proposed method provided better results in all three applications than the existing methods. Chen et al. (2014) presented a new method for group decision making using incomplete fuzzy preference relations based on the additive consistency and the order consistency with consistency degrees.

Different types of FSs are used to represent the words or linguistic information, elicited by human beings. To process such linguistic information, Zadeh put forward the paradigm of Computing with words (CWW) (Zadeh 1996). CWW is a remarkable mathematical technique that aims to process the linguistic information to generate recommendations. There are different types of CWW approaches viz., extension principle, symbolic method, 2-tuple approach and perceptual computing (Per-C). The differentiating parameter for all the methodologies is principally the way in which they represent and process the semantics of linguistic information, through FSs. CWW methodologies have been used in a number of applications.

One such application area is the student learning assessment as well as examination evaluation. Huyapaya (2012) developed a methodology, which analyzed the overall learning capability of a student using the fuzzy logic systems. The methodology considered various uncertainties associated with the system such as student's behavior, student's rate of learning and improvement, etc. However, we feel that the methodology left out other important parameters such as student's motivation and the difficulty of the subject, while performing the analysis. Saleh and Kim (2009) proposed a system for evaluating the student performance using the concepts of fuzzy logic. Hameed and Sorensen (2010) improved the system of (Saleh and Kim 2009), for student performance evaluation. In this paper, triangular membership functions (MFs) used in fuzzy logic system of (Saleh and Kim 2009) were replaced by the Gaussian MFs. This optimized the student performance scores, according to the authors. Biswas (1995) presented a useful method to assess the students' answer scripts.

Sevarac (2006) presented a system where the students and teachers interacted using the linguistic variables and the FSs. The system used the neuro-fuzzy mathematical technique. Students' behavior formed the key factor for use in system modeling. Aye and Thwin (2008) presented various advantages of conducting examinations using the smartphones and the limitations in the implementation of the same. Gupta (2012) attempted to overcome these limitations thereby establishing the fact that mobile examination systems were the new direction for the conduct of examinations. However, the system proposed in (Gupta 2012) was implemented on Nokia Qt SDK for Nokia-Symbian operating system enabled smartphones, which restricted its wider applicability as android was an open source software. So later, the system was implemented on android enabled devices (Gupta et al. 2014). Wang and Chen (2008) proposed a remarkable paper for evaluating students' answer scripts using fuzzy numbers associated with confidence of the evaluator. The method could evaluate students' answer scripts in a more flexible and intelligent manner.

In all these works involving the evaluation of students' answer scripts, the primary objective was to assess the subject knowledge of any student, which he/ she acquired while studying a subject[3]. The various parameters used to judge the student's performance took linguistic values. The students undertook the examination (of a subject) and answered the questions. The number of questions that they solved accurately judged their success rate in the examinations.

---

[3] The subjects are taught as a part of curriculum of a course like bachelors, masters, etc., that a student is enrolled in.

However, we feel that, only answering the questions correctly is not a true test of a student's ability. If a student solves any question faster than the other students do, then that student should get more marks for that question. This is because the systematic solution methodology[4] adopted by the student is better than that of other students. We have referred to this question solving methodology of the student as the student's strategy[5]. Student strategy evaluation is the subjective judgment about the student's strategy as perceived by the faculty member, who is evaluating the students' answer scripts. Thus, this subjective nature of judgment motivated us to use the CWW paradigm for student strategy evaluation.

The main contribution of this paper is to illustrate the use of CWW for student strategy evaluation and present a comparison of the recommendations generated by different CWW approaches viz., extension principle, symbolic method, 2-tuple and Per-C. CWW provides us with two major advantages. Firstly, it generates a numeric score for the overall evaluation of strategy adopted by a student in the examination. This enables comparison and ranking of the students based on their performances. Secondly, a linguistic evaluation describing the student strategy is also obtained from the system. Both these numeric score and linguistic recommendation corresponding to the student's strategy are together taken as a measure of his/ her learning outcome. The goal of assessing the learning outcome is to ensure that all the students, especially the ones who have low academic learning, are paid more attention by the faculty members.

Furthermore, the linguistic recommendation is useful for human beings as they naturally understand and express themselves using 'words', 'words' being treated as fuzzy information granules in the GC paradigm, which is perhaps the case with most of the human reasoning and concepts. Also, through the comparison of the recommendations generated by different CWW approaches, we found that Per-C outperforms the others CWW approaches by generating unique recommendations in 100% cases as well as modeling the word uncertainty in the best possible way, which other approaches failed to do. The main outcome of this paper is a novel Per-C based approach for student strategy evaluation, which to the best of our knowledge has not been proposed before.

Rest of the paper is organized as follows: Section 2 provides an introduction of FSs and CWW, Section 3 provides a description of the extension principle, symbolic method, and the 2-tuple based CWW approach. It also details the methodology for using these three approaches for student strategy evaluation. Section 4 describes the mathematical details of perceptual computing as well as describes how to use it for student strategy evaluation. Section 5 compares the recommendations obtained by application of various CWW approaches for strategy evaluation considering the test case of 25 students. Finally, Section 6 concludes the paper and highlights its future work.

## 2 Fuzzy Sets and CWW

Linguistic data or 'words' (Zadeh 1975) are uncertain because 'words mean different things to different people'. For example, we conducted a survey among a group of students, based on the data taken from a website that depicted the university rankings of a country at global level. To the best of our knowledge, the performance of the universities were given in 5 star ratings and the highest rating achieved by a university was 4.5. When we asked the students whether they considered 4.5 to be a 'good' or 'bad' rating, we found that students gave varying feedback between these two extremes viz. students considered rating to be 'very good', 'moderately good', 'fair', etc. Though the overall rating is precise numeric value of 4.5, every user has a subjective interpretation for the same thus depicting the inherent uncertainty in the linguistic data.

Therefore, Zadeh (1965) proposed the concept of FSs to capture and model the uncertainty of linguistic data in best possible way. FSs are an extension of ordinary or 'crisp' sets. Crisp sets place a full degree of affirmation on the belongingness of an object to a set. Consider an example from (Zadeh 1965). Dogs, cats, etc. will unambiguously belong to a crisp set defined as collection of animals. Plants, rocks, etc. will definitely not belong to that set. Therefore, an object is either completely included or completely excluded from a crisp set. The degree of belongingness of an object to a set is also called its MF. In case of crisp sets, the MF of an object is either '1' or '0', '1' being the belongingness to the set and '0' being the exclusion from the set. MF of an object in a crisp set cannot take any intermediate values between these two extremes of '1' and '0'. However, consider the above example again for set of animals. As stated in (Zadeh 1965), the entities like starfish have an ambiguous status with respect to the belongingness to the crisp set of animals. One cannot say with complete affirmation that starfish is an animal because some people consider it as an animal, some don't consider it as an animal, some consider it partially as an animal, etc. In such scenarios, the collection of objects is represented as FSs. FSs allow the MF of an object to take any value from the closed interval [0, 1].

---

[4] By solution methodology, we mean the collection of number of steps that form the solution of the question.

[5] A shorter version of the present work has been presented at UKSIM 2015 (Gupta et al. 2015). A number of similar works in support of our claim can be found in (Gupta 2012; Sripan and Suksawat 2010; Hameed and Sorensen 2010; Sevarac 2006).

Every element belonging to a FS is expressed mathematically as a combination of two terms: $(x, \mu_x)$ where $x$ is the element belonging to the set and $\mu_x$ is its MF. Therefore, in case of FSs, $\mu_x \in [0,1]$ and crisp sets, $\mu_x \in \{0,1\}$. The FSs proposed by Zadeh in (Zadeh 1965), are called the type-1 (T1) FSs. These sets represent the MF values as crisp or precise values.

Later it was realized that the membership value of a FS cannot be precise because FSs are used for representation of uncertain quantities. The representation of an uncertain quantity cannot be a precise value. So, Zadeh himself proposed the idea of higher-order fuzzy sets of which the most commonly used are the type-2 (T2) FSs[6] (Zadeh 1975). T2 FSs provide additional information for each set element, the uncertainty about the MF value and thus extend the concept of T1 FSs. Zadeh (1975) also conceptualized a special T2 FS, in which this uncertainty about the membership value are assumed as 1 for all set elements and called such FSs as interval type 2 fuzzy sets (IT2 FSs). Karnik, Mendel and Liang proposed an IT2 fuzzy logic system using the concept of IT2 FSs in 1999 (Karnik et al. 1999). IT2 FSs have an enhanced capability to capture and model the 'word' uncertainty in a better manner than the T1 FSs (Mendel and Wu 2010).

Mathematically, the T2 FSs are denoted as $(x, \mu_x, J_x)$, where the third variable $J_x$ is added to the T1 FS representation and denotes the uncertainty about $\mu_x$. In T2 FSs, $\mu_x$ is called the primary MF and the $J_x$ is called secondary MF. If $J_x = 1$ everywhere then T2 FSs reduce to IT2 FSs. Graphically, an IT2 FS is shown in Fig. 1.

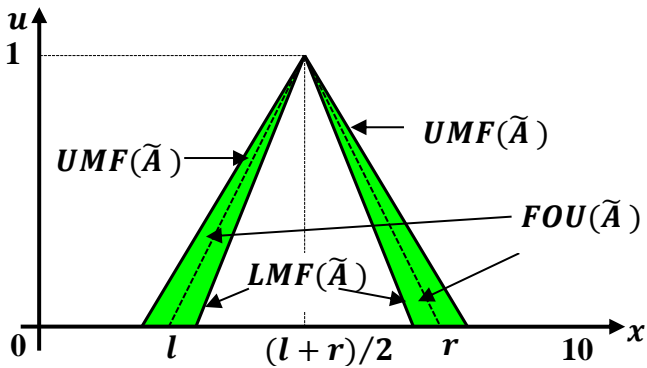

Fig. 1: Graphical representation of IT2 FS

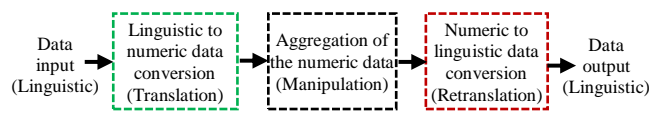

Fig 2: CWW methodology (Yager 1999; Mendel 2007)

[6] Some literature on type-2 FSs may also be found in (Mendel 2003; Greenfield and John 2009)

In Fig. 1, the quantity on the y-axis is the membership value ($u$) and on the x-axis is the variable. The UMF stands for upper membership function and LMF stands for lower membership function. Both the UMF and LMF are T1 FSs. The shaded region in the Fig. is called the Footprint of Uncertainty (FOU). The FOU is bounded from above by the UMF and below by the LMF.

In the IT2 FSs, since the secondary membership value is 1 everywhere, therefore the spread of FOU gives an idea of the uncertainty captured by the IT2 FS. The FOU can be visualized to be sitting atop the primary MF of the word that the IT2 FS models. There is a dashed line or T1 FS depicted in Fig. 2, whose left and right ends rest on the x-axis at $l$ and $r$, respectivley. It's centre at $\dfrac{(l+r)}{2}$ has a membership value of 1. This T1 FS is called an embedded T1 FS. The FOU can be assumed to be a union of continuum of such embedded T1 FSs.

To process the linguistic information represented as FSs, Zadeh proposed the new paradigm of CWW (Zadeh 1996) and thus coined the term CWW. CWW can facilitate machines to receive human perceptions as input (which are generally expressed as words), manipulate them and generate recommendations. The basic principle underlying a CWW methodology is the one to one mapping between the linguistic and the numeric information. CWW is a methodology in which words and propositions drawn from natural language are the objects of computation. For example 'small', 'large', 'far', 'heavy', 'not very likely', 'extremely', etc. Any system based on CWW performs three basic steps shown in Fig. 2 and is quite useful whenever the linguistic information needs to be processed using a computer. The data input and output to the CWW system are both in linguistic form. Fig. 2 is Yager's representation of CWW (Yager 1999; Mendel 2007). It can be seen that, in Yager's CWW methodology, there are three building blocks: translation, manipulation and retranslation.

These words activate the CWW system, which converts these words into mathematical representation using the FSs in the first step of translation. This is required because computers understand the numeric information and not the linguistic one. The input information to a CWW system may be from different sources or a single source may provide multiple pieces of information. So, it needs to be aggregated and used for generation of recommendations. This is done in the next step of manipulation. The final step is to convert the aggregated numeric information back to the linguistic form because humans naturally prefer linguistic information. Retranslation step does this.

There are different types of CWW methodologies based on this Yager's model. The differentiating parameter for all the methodologies is the way in which the semantics of linguistic information is represented. A popular CWW methodology, that represents the semantics of linguistic information using T-1 FSs, is the extension principle based CWW methodology. It operates on the MF of the linguistic terms. Another famous CWW methodology is the symbolic method, which operates on the indices of linguistic terms contained in the linguistic term set. Both the extension principle and symbolic method fail to give unique recommendations in certain situations, which was illustrated by Herrera and Martinez (2000). Therefore, they proposed a novel CWW methodology called the 2-tuple fuzzy linguistic approach, which combines the advantages of both the extension principle and symbolic method. It operates on the indices of the linguistic terms and represents the semantics of linguistic terms using T-1 FS MFs of triangular shape.

However, we found that the 2-tuple approach also suffers from some drawbacks. Firstly, it represents the semantics of linguistic terms using T1 FSs. T1 FSs do not capture and model the 'word' uncertainty in the best possible manner (Castillo and Melin 2008). Therefore, Mendel et al. proposed a novel CWW methodology, called 'Perceptual Computing (Per-C)', which represents the 'word' semantics using IT2 FSs (Mendel 2007). When compared with Yager's CWW methodology (shown in Fig. 2), the corresponding three building blocks of Per-C are Encoder, CWW engine and Decoder, respectively. Per-C was used in a number of real-life applications such as investment judgment analysis (Mendel and Wu 2010), power optimization of battery operated devices (Gupta and Muhuri 2014; Gupta and Muhuri 2016; Muhuri et al. 2017), perceptual reasoning (Mendel and Wu 2008), fuzzy control applications (Cazarez-Castro et al. 2012, Castro 2011), etc.

## 3 Student strategy evaluation using Extension principle, Symbolic method and 2-tuple based CWW methodologies

In this section we will discuss the details of extension principle, symbolic method and 2-tuple based CWW approaches (Herrera and Martinez 2000) and use them for assessing a student's strategy in an examination.

Whenever the students, sit the examination of any subject, they are presented with different questions on varying levels of difficulty. They answer the questions to the best of their ability, using their respective solution methodologies or strategies. We identified certain parameters that can be used to assess the quality of the respective student's strategy viz., Time taken to solve the question, Subject's Knowledge, Liking towards Subject and Perceived preparation level. These parameters along with their linguistic values, are listed in Table 1.

**Table 1** Parameters and their linguistic values

| | Parameter/ Recommendation | Linguistic values |
|---|---|---|
| Parameter | Time taken to solve the question | Very little (VL) |
| | | Small (S) |
| | | Moderate (M) |
| | | Large (L) |
| | | Very Large (VLA) |
| | Subject's Knowledge | Very Limited (SVL) |
| | | Limited (SL) |
| | | Moderate (SM) |
| | | Large (SLA) |
| | | Very Large (SVLA) |
| | Liking towards Subject | Very Less (AVL) |
| | | Less (AL) |
| | | Moderate (AM) |
| | | High (AH) |
| | | Very High (AVH) |
| | Perceived preparation level | Very Less (PVL) |
| | | Less (PL) |
| | | Moderate (PM) |
| | | High (PH) |
| | | Very High (PVH) |
| Recommendation | Strategy of student | Not Good (SSNG) |
| | | Below Average (SSBA) |
| | | Average (SSA) |
| | | Good (SSG) |
| | | Very Good (SSVG) |

Feedback about each parameter is provided using exactly one of the corresponding linguistic values. For example, Time taken to solve the question can take exactly one values out of 'Very little', 'Small', 'Moderate', 'Large' or 'Very Large'. Similar is the case for other parameters. Then using a CWW approach recommendations are generated regarding the quality of strategy. Each of the CWW methodologies generates a strategy score, which can be used to compare and rank the respective students based on their strategies. In addition, each CWW methodology utilizes this strategy score to generate a linguistic recommendation describing the student strategy.

It is mentioned here that an initial work in the direction was (Gupta et al. 2015). Inspired from this work, we modified some of the parameters and the linguistic values used for the assessment of student's strategy. We also dropped certain parameters such as 'stability before the examination', 'planning', 'subject phobia' and the 'expectation'. We changed the name of the parameter 'practice' to 'perceived preparation level'. Our aim was to improve the student strategy assessment model.

## 3.1 Extension principle based student strategy evaluation

The extension principle represents the linguistic terms of the term set in the form of triangular T1 MF and each term as a collection of tri-tuple: $(l, m, r)$, $l$ being the left end of the triangle, $m$ it's middle and $r$ the right end. Computations are performed on these tri-tuples and recommendations are generated.

Consider a set of linguistic terms $S = \{s_0, s_1, \ldots \ldots, s_g\}$. The cardinality of set is $g + 1$. Each term is represented by a triangular membership function centered at $i/g$, $i = 0 \, to \, g$. For example, in a real life problem, we are representing the linguistic variable '$temperature$' that can take linguistic values '$very \, cold \, (vc)$', '$cold \, (c)$', '$moderate \, (m)$', '$hot \, (h)$' and '$very \, hot \, (vh)$'. Other linguistic values are also possible, but we have chosen only these five for simplicity of discussion. They are represented by triangular MFs and are shown in Fig. 3, with each triangular T1 MFs centered at $i/4$, $i = 0 \, to \, 4$.

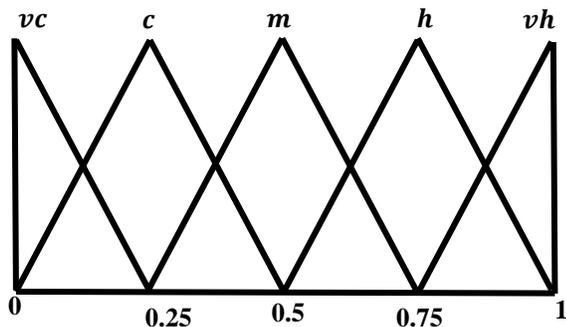

Fig 3: Triangular membership function representation of 'temperature'

### Algorithm 1: Extension principle based CWW

1. Decide a linguistic term set $S = \{s_0, s_1, \ldots \ldots, s_g\}$. The cardinality of set is $g + 1$, corresponding to the problem.
2. Different stakeholders provide their prefernces using the linguistic terms drawn from $S$.
3. Let there be $i$ number of linguistic preferences or pieces of information, each of which is drawn from $S$.
4. Denote the collective preference vector as:

$$\{s_{j1}, s_{j2}, \ldots s_{ji}\}$$

where $j = 0 \, to \, n - 1$ and each $s_{jk} \in S; k = 1 \, to \, i$.

5. Represent each of these linguistic preferences by a tri-tuple $(l_{ji}, m_{ji}, r_{ji})$, where each value in the tuple corresponds to the left, middle and right end, respectively of the triangular membership function. Therefore, preference vector becomes:

$$\{(l_{j1}, m_{j1}, r_{j1}), (l_{j2}, m_{j2}, r_{j2}), \ldots (l_{ji}, m_{ji}, r_{ji})\}$$

6. Aggregate these tri-tuples to obtain the collective performance vector $C$ as:

$$C = (l_c, m_c, r_c) = \left( \frac{1}{i}\sum_{k=1}^{i} l_{jk}, \frac{1}{i}\sum_{k=1}^{i} m_{jk}, \frac{1}{i}\sum_{k=1}^{i} r_{jk} \right)$$

7. Perform linguistic approximation to generate a linguistic output. Calculate the Euclidian distance of the performance vector from each of the linguistic terms $s_p = (l_q, m_q, r_q) \in S, q = 1 \, to \, n - 1$, in tri-tuple form as:

$$d(s_q, C) = \sqrt{P_1(l_q - l_c)^2 + P_2(m_q - m_c)^2 + P_3(r_q - r_c)^2}$$

where the $P_i, i = 1,2,3$ are the weights, with values 0.2, 0.6 and 0.2 respectively.

8. The recommended linguistic term is $s_b^* \in S$, such that $d(s_b^*, C) \leq d(s_q, C), \forall s_q \in S$, viz., one with maximum similarity or minimum distance.

Consider a scenario where different stakeholders are providing their linguistic preferences. Let there be $i$ number of linguistic preferences or pieces of information, each of which is drawn from $S$. Let the collective preference vector is denoted as Eq. (1):

$$\{s_{j1}, s_{j2}, \ldots s_{ji}\} \tag{1}$$

where $j = 0 \, to \, n - 1$ and each $s_{jk} \in S; k = 1 \, to \, i$. In the first step, the extension principle represents each of these linguistic preferences by a tri-tuple $(l_{ji}, m_{ji}, r_{ji})$, where each value in the tuple corresponds to the left, middle and right end, respectively of the triangular membership function. Therefore, Eq. (1) becomes:

$$\{(l_{j1}, m_{j1}, r_{j1}), (l_{j2}, m_{j2}, r_{j2}), \ldots (l_{ji}, m_{ji}, r_{ji})\} \quad (2)$$

In the next step, these tri-tuples are aggregated to obtain the collective performance vector $C$, which is also a tri-tuple. The calculation of $C$ is done as:

$$C = (l_c, m_c, r_c) = \left(\frac{1}{i}\sum_{k=1}^{i} l_{jk}, \frac{1}{i}\sum_{k=1}^{i} m_{jk}, \frac{1}{i}\sum_{k=1}^{i} r_{jk}\right) \quad (3)$$

The collective performance vector, Eq. (3), does not usually match any linguistic terms in the term set $S$. Therefore, to generate a linguistic output, the process of linguistic approximation is used. In the process of linguistic approximation, the Euclidian distance of the performance vector is calculated from each of the linguistic terms $s_p = (l_q, m_q, r_q) \in S, q = 1 \ to \ n-1$, in tri-tuple form. The term with maximum similarity or minimum distance is recommend as the output. This is denoted in Eq. (4) as:

$$d(s_q, C) = \sqrt{P_1(l_q - l_c)^2 + P_2(m_q - m_c)^2 + P_3(r_q - r_c)^2} \quad (4)$$

where the $P_i, i = 1,2,3$ are the weights, with values 0.2, 0.6 and 0.2 respectively. The recommended linguistic term is $s_b^* \in S$, such that $d(s_b^*, C) \le d(s_q, C), \forall s_q \in S$. We summarize the working of extension principle based CWW approach in the form of Algorithm 1.

Now we use the extension principle for student strategy evaluation. The parameters used to evaluate the student's performance are given in Table 1. We define the linguistic term sets corresponding to these parameters as in Eq. (5):

*Time taken to solve the question*:
$\{s_0: vl, s_1: s, s_2: m, s_3: l, s_4: vla\}$

*Subject's Knowledge*:
$\{s_0: svl, s_1: sl, s_2: sm, s_3: sla, s_4: svla\}$

*Liking towards subject*:
$\{s_0: avl, s_1: al, s_2: am, s_3: ah, s_4: avh\}$

*Perceived preparation level*:
$\{s_0: pvl, s_1: pl, s_2: pm, s_3: ph, s_4: pvh\}$

*Strategy of student*:
$\{s_0: ssng, s_1: ssba, s_2: ssa, s_3: ssg, s_4: ssvg\}$ (5)

The expanded forms of the linguistic terms given in Eq. (5), can be seen from Table 1. We represent the linguistic terms of each parameter uniformly on scale of 0 to 1 in the form of triangular MFs, shown in Fig. 4.

Consider the feedback of student SS1 for various parameters shown in Table 2. Thus, collective preference vector for him/ her, similar to Eq. (1) is given by Eq. (6) as:

*Performance of SS1* $= \{s, sla, am, pm\}$ (6)

**Table 2** Feedback of Two Students

| Parameter | Linguistic feedback of two students | |
|---|---|---|
| | Student 1 (SS1) | Student 2 (SS2) |
| Time taken to solve the question | Small | Large |
| Subject's Knowledge | Large | Limited |
| Liking towards Subject | Moderate | High |
| Perceived preparation level | Moderate | Less |

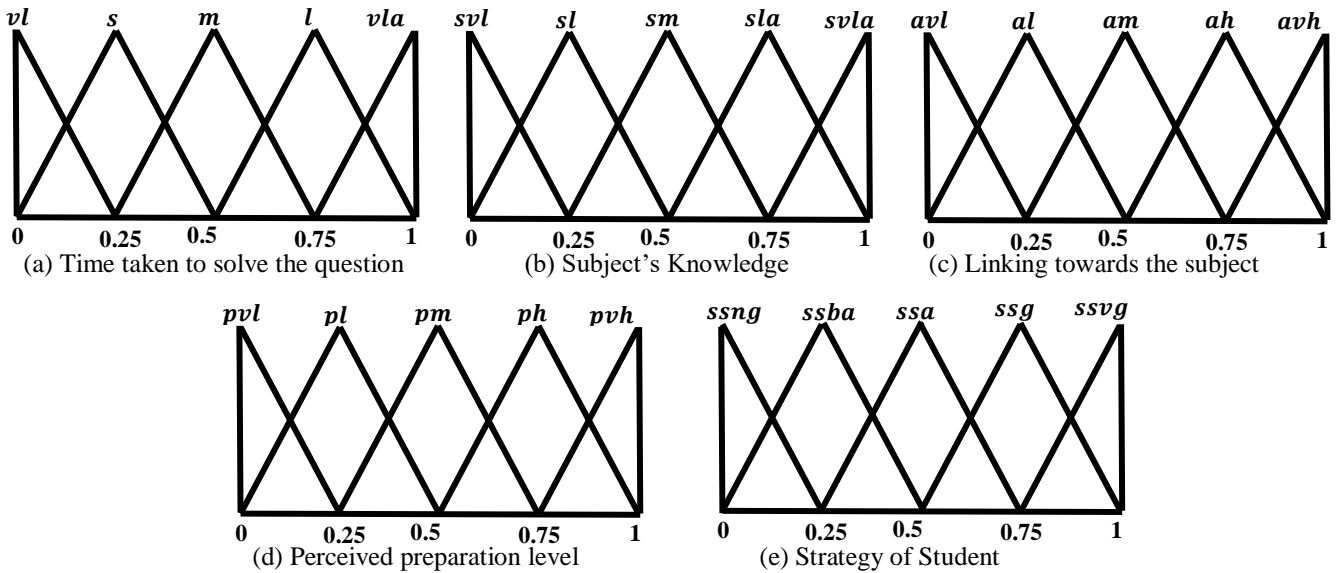

(a) Time taken to solve the question
(b) Subject's Knowledge
(c) Linking towards the subject
(d) Perceived preparation level
(e) Strategy of Student

Fig. 4: Triangular MF representation of linguistic terms of parameters

**Table 3** End Points of Linguistic Feedback of Two Students

| Parameter | Feedback of students | | | |
| --- | --- | --- | --- | --- |
| | Student 1 (SS1) | | Student (SS2) | |
| | Linguistic feedback | Tri-tuple end points | Linguistic feedback | Tri-tuple end points |
| Time taken to solve the question | Small | $\{0, 0.25, 0.5\}$ | Large | $\{0.5, 0.75, 1\}$ |
| Subject's Knowledge | Large | $\{0.5, 0.75, 1\}$ | Limited | $\{0, 0.25, 0.5\}$ |
| Liking towards Subject | Moderate | $\{0.25, 0.5, 0.75\}$ | High | $\{0.5, 0.75, 1\}$ |
| Perceived preparation level | Moderate | $\{0.25, 0.5, 0.75\}$ | Less | $\{0, 0.25, 0.5\}$ |

Using Fig. 4, the ends of the triangular MFs corresponding to the feedback of SS1 are found, similar to Eq. (2). Thus, feedback of student is represented in the tri-tuple form is given in Eq. (7) as:

$$\{(0, 0.25, 0.5), (0.5, 0.75, 1), (0.25, 0.5, 0.75), (0.25, 0.5, 0.75)\} \quad (7)$$

The collective performance vector for the student SS1 is calculated using Eq. (3) and is found to be:

$$C = \{0.25, 0.5, 0.75\} \quad (8)$$

To generate the linguistic recommendation, we perform computations, similar to Eq. (4) between the collective performance vector from Eq. (8) and the term set '*strategy of student*' from Eq. (5). Thus, the recommended output is that the strategy of student is '*ssa*' or '*average*'. Proceeding similarly for student (SS2), the performance is also found to be '*average*'.

Table 3 lists the tri-tuple end point values corresponding to the feedback of students SS1 and SS2. For example, the time taken to solve the question by SS1 is '*small*', whose triangular MF is characterized by tri-tuple end points $\{0, 0.25, 0.5\}$, as shown in Fig. 4(a). Therefore, the value is depicted as $\{0, 0.25, 0.5\}$, in fourth row and third column. Similarly, other values in the Table 3 are interpreted.

### 3.2 Symbolic method for student strategy evaluation

Symbolic method operates on the indices of linguistic terms in the term set. Consider the linguistic preference set from Eq. (1), $\{s_{j1}, s_{j2}, \dots \dots s_{ji}\}$. The symbolic method assigns a weight from the weight vector $W = [w_1, \dots \dots \dots \dots w_i]$; corresponding to each of the $ith$ linguistic information term in the linguistic preference set; such that each of $w_p \in [0,1]; p = 1 \text{ to } i$ and $\sum_{p=1}^{i} w_p = 1$. The technique first orders the linguistic term set containing the linguistic feedback and then aggregates the data values according to the function $(SM^j)$ given in Eqs. (9)-(10):

For j>2,
$$SM^j\{w_k, s_k, k = 1, .. j\} = (w_1 \odot s_1) \oplus \left((1 - w_1) \odot SM^{j-1}\{\sqcap_h, s_h, h = 2, .. j\}\right) \quad (9)$$

where $\sqcap_h = {}^{w_h}\!\big/_{\sum_{l=2}^{j} w_l}; h = 2, 3, \dots \dots \dots j$

For j=2,
$$SM^2\{\{w_1, 1 - w_1\}, \{s_l, s_q\}\} = (w_1 \odot s_l) \oplus (1 - w_1 \odot s_q) = s_r \quad (10)$$

such that $r = \min\{g, q + round(w_1.(q - l))\}$; $g + 1$ being the cardinality of the set to be aggregated and $round()$ is the usual round function.

---

**Algorithm 2: Symbolic method based CWW**

1. Decide a linguistic term set $S = \{s_0, s_1, \dots \dots \dots, s_g\}$. The cardinality of set is $g + 1$, corresponding to the problem. Different stakeholders provide their preferences using the linguistic terms drawn from $S$.

2. Let there be $i$ number of linguistic preferences or pieces of information, each of which is drawn from $S$.
   Denote the collective preference vector as:
   $$\{s_{j1}, s_{j2}, \dots s_{ji}\}$$
   where $j = 0 \text{ to } n - 1$ and each $s_{jk} \in S; k = 1 \text{ to } i$.

3. Order the linguistic term set of Eq. (1) in increasing order of indices of linguistic terms and assign a weight from the weight vector $W = [w_1 \dots \dots \dots \dots w_i]$; corresponding to each of the $ith$ information term; such that each of $w_p \in [0,1]; p = 1 \text{ to } i$ and $\sum_{p=1}^{i} w_p = 1$.

4. Aggregate the data values as:
   For j>2,
   $$SM^j\{w_k, s_k, k = 1, .. j\} = (w_1 \odot s_1) \oplus \left((1 - w_1) \odot SM^{j-1}\{\sqcap_h, s_h, h = 2, .. j\}\right)$$
   where $\sqcap_h = {}^{w_h}\!\big/_{\sum_{l=2}^{j} w_l}; h = 2, 3, \dots \dots \dots j$
   For j=2,
   $$SM^2\{\{w_1, 1 - w_1\}, \{s_l, s_q\}\} = (w_1 \odot s_l) \oplus (1 - w_1 \odot s_q) = s_r$$
   such that $r = \min\{g, q + round(w_1.(q - l))\}$; $g + 1$ being the cardinality of the set to be aggregated and $round()$ is the usual round function.

5. $s_r$ is the recommended linguistic term and $r$ is the numerical value.

---

**Table 4** Linguistic Feedback of Two Students in Symbolic Method Representation

| Parameter | Feedback of students | | | |
|---|---|---|---|---|
| | Student 1 (SS1) | | Student (SS2) | |
| | Linguistic feedback | Symbolic method representation | Linguistic feedback | Symbolic method representation |
| Time taken to solve the question | Small | $s_1$ | Large | $s_3$ |
| Subject's Knowledge | Large | $s_3$ | Limited | $s_1$ |
| Liking towards Subject | Moderate | $s_2$ | High | $s_3$ |
| Perceived preparation level | Moderate | $s_2$ | Less | $s_1$ |

It can be seen that aggregation function $(SM^j)$ performs convex combination of information at each step proceeding in top-down manner. At each step of the aggregation, computations are performed on the numeric indexes of the term set to give an aggregated value as $s_r$. Finally, the recommended value is a unique index of the term belonging to the term set $S$. The overall working of the symbolic method based CWW is shown in the form of Algorithm 2.

Now we use the symbolic method for student strategy evaluation. The parameters used to evaluate the student's performance are given in Table 1 and the term set corresponding to these parameters is given in Eq. (5).

Consider again the feedback of student SS1 for various parameters shown in Table 2. Using the Eq. (5), the collective preference vector for SS1 is given as:

$$Performance\ of\ SS1 = \{s_1: s, s_3: sla, s_2: am, s_2: pm\} \quad (11)$$

These linguistic values are first ordered according to the indices of the linguistic terms. Thus, Eq. (11) becomes Eq. (12) as:

$$Performance\ of\ SS1 = \{s_3, s_2, s_2, s_1\} \quad (12)$$

We use the weight matrix as:

$$W = \left[w_1 = \frac{1}{4}, w_2 = \frac{1}{4}, w_3 = \frac{1}{4}, w_4 = \frac{1}{4}\right] \quad (13)$$

For aggregation we perform computations similar to Eqs. (9)-(10) on data values of Eq. (12) and weight matrix in Eq.(13). These are shown step-by-step with explanations in Eqs. (14)-(18) as:

$$SM^4\left\{\left[w_1 = \frac{1}{4}, w_2 = \frac{1}{4}, w_3 = \frac{1}{4}, w_4 = \frac{1}{4}\right], [s_3, s_2, s_2, s_1]\right\}$$
$$= \left(\frac{1}{4} \odot s_3\right) \oplus \left(\frac{3}{4} \odot SM^3\left\{\left[\frac{1}{3}, \frac{1}{3}, \frac{1}{3}\right], [s_2, s_2, s_1]\right\}\right) \quad (14)$$

$$SM^3\left\{\left[\frac{1}{3}, \frac{1}{3}, \frac{1}{3}\right], [s_2, s_2, s_1]\right\}$$
$$= \left(\frac{1}{3} \odot s_2\right) \oplus \left(\frac{2}{3} \odot SM^2\left\{\left[\frac{1}{2}, \frac{1}{2}\right], [s_2, s_1]\right\}\right) \quad (15)$$

$$SM^2\left\{\left\{\frac{1}{2}, \frac{1}{2}\right\}, \{s_2, s_1\}\right\} = \left(\frac{1}{2} \odot s_2\right) \oplus \left(\frac{1}{2} \odot s_1\right) = s_r \quad (16)$$

In Eq. (16), $r = \min\left(4, 1 + round(\frac{1}{2} * (2 - 1))\right) = \min(4, 2) = 2$. Therefore, result of Eq. (16) is $s_2$. Substituting $s_2$ in Eq. (15), we get Eq. (17) as:

$$SM^3\left\{\left[\frac{1}{3}, \frac{1}{3}, \frac{1}{3}\right], [s_2, s_2, s_1]\right\} = \left(\frac{1}{3} \odot s_2\right) \oplus \left(\frac{2}{3} \odot s_2\right) = s_r \quad (17)$$

In Eq. (17), $r = \min\left(4, 2 + round(\frac{1}{3} * (2 - 2))\right) = \min(4, 2) = 2$. Therefore, result of Eq. (17) is $s_2$. Substituting $s_2$ in Eq. (14), we get Eq. (18) as:

$$SM^4\left\{\left[w_1 = \frac{1}{4}, w_2 = \frac{1}{4}, w_3 = \frac{1}{4}, w_4 = \frac{1}{4}\right], [s_3, s_2, s_2, s_1]\right\}$$
$$= \left(\frac{1}{4} \odot s_3\right) \oplus \left(\frac{3}{4} \odot s_2\right) = s_r \quad (18)$$

In Eq. (18), $r = \min\left(4, 2 + round(\frac{1}{4} * (3 - 2))\right) = \min(4, 2) = 2$. Therefore, result of Eq. (18) is $s_2$.

Therefore, the recommended linguistic term corresponding to strategy of student is '$ssa$' or '$average$'. Proceeding similarly for student (SS2), the performance is also found to be '$average$'.

Table 4 lists the indices corresponding to the linguistic feedback of students SS1 and SS2. For example, the time taken to solve the question by SS1 is '$small$', which occurs at index 1 $(s_1)$ in the term set $Time\ taken\ to\ solve\ the\ question$ given in Eq. (5). Therefore, the value is depicted as $s_1$, in fourth row and third column. Similarly, other values in the Table 4 are interpreted.

### 3.3 2-tuple based CWW methodology for student strategy evaluation

2-tuple approach for CWW is inspired from both the extension principle and the symbolic method. In 2-tuple approach, each information is represented as a twin value $(s, \alpha)$, where $s \in S$, is a term drawn from the term set $S$ and $\alpha \in [-0.5, 0.5)$ is called the symbolic translation. In any decision making problem, let the result of aggregation of preferences obtained from multiple experts be $\beta$. The recommended solution is given by $(s_{round(\beta)}, \alpha)$ where, $\alpha = \beta - round(\beta)$. Here $round$ is the usual mathematical rounding operation. For example, the result of aggregation in a decision problem is 2.3. So, $\beta = 2.3$, $round(2.3) = 2, \alpha = 2.3 - 2 = 0.3$. The solution is given by $(s_2, 0.3)$.

We now explain the CWW approach based on 2-tuple. Consider the linguistic preference term set of the stakeholders given in the Eq. (1). viz., $\{s_{j1}, s_{j2}, \ldots \ldots, s_{ji}\}$. To process the linguistic preferences in this set using 2-tuple approach, in the first step every element of this set is convert into twin tuple. Therefore, the resulting set becomes:

$$\{(s_{j1}, \alpha_1), (s_{j2}, \alpha_2), \ldots \ldots (s_{ji}, \alpha_i)\} \quad (19)$$

As each of $s_{jk} \in S, k = 1 \, to \, i$, therefore all the $\alpha_k = 0$ (Herrera and Martinez 2000). We obtain a value $\beta_{2tp} \in [0, g], g + 1$ being the cardinality of the set; by performing aggregation on the indices of these linguistic terms using arithmetic mean as:

$$\beta_{2tp} = \frac{j1 + j2 + \cdots + ji}{i} \quad (20)$$

The symbolic translation $\alpha_{2tp}$ is then obtained as:

$$\alpha_{2tp} = \beta_{2tp} - round(\beta_{2tp}), \alpha_{2tp} \in [-0.5, 0.5) \quad (21)$$

Thus, the recommended linguistic information is:

$$\left(s_{round(\beta_{2tp})}, \alpha_{2tp}\right) \quad (22)$$

The working of 2-tuple based CWW approach is summarized in the form of Algorithm 3.

Now we will illustrate the student strategy evaluation using the 2-tuple approach. Consider feedback of Student (SS1), given in Table 2. The indices of various linguistic values for the parameters are given in Eq. (11). These linguistic values are converted to 2-tuple form, similar to Eq. (19). Since each linguistic value is directly drawn from the term set, therefore, each one has the translation distance 0. Therefore, the preference vector becomes:

$$\{(s_1, 0), (s_3, 0), (s_2, 0), (s_2, 0)\} \quad (23)$$

The preference vector of Eq. (23) is aggregated using Eq. (20) as:

$$\beta_{2tp} = \frac{1 + 3 + 2 + 2}{4} = 2 \quad (24)$$

The translation distance is calculated using Eq. (21) as:

$$\alpha_{2tp} = \beta_{2tp} - round(\beta_{2tp}) = round(2) - 2 = 0 \quad (25)$$

Therefore, the recommended linguistic term is found using Eq. (22) as:

$$\left(s_{round(\beta_{2tp})}, \alpha_{2tp}\right) = (s_2, 0) = (ssa, 0) \quad (26)$$

Therefore, the recommended linguistic term is given as: $(ssa, 0)$ or '$average$'. Proceeding similarly for student (SS2), the performance is also found to be '$average$'.

Table 5 gives the feedback of students SS1 and SS2 in 2-tuple form. For example, the time taken to solve the question by SS1 is '$small$', which occurs at index 1 in the term set $Time\ taken\ to\ solve\ the\ question$ given in Eq. (5). Also, its translation distance is 0. Therefore, the value is depicted as $(s_1, 0)$, in fourth row and third column. Similarly, other values in the Table 5 are interpreted.

| Algorithm 3: 2-Tuple based CWW |
|---|

1. Decide a linguistic term set $S = \{s_0, s_1, \ldots \ldots, s_g\}$. The cardinality of set is $g + 1$, corresponding to the problem.
2. Different stakeholders provide their prefernces using the linguistic terms drawn from $S$.
3. Let there be $i$ number of linguistic preferences or pieces of information, each of which is drawn from $S$.
4. Denote the collective preference vector as:

$$\{s_{j1}, s_{j2}, \ldots s_{ji}\}$$

where $j = 0 \, to \, n - 1$ and each $s_{jk} \in S; k = 1 \, to \, i$.

5. Convert the linguistic preference vector of Eq. (1) to 2-tuple form as:

$$\{(s_{j1}, \alpha_1), (s_{j2}, \alpha_2), \ldots \ldots (s_{ji}, \alpha_i)\}$$

6. Aggregate these data values as:

$$\beta_{2tp} = \frac{j1 + j2 + \cdots + ji}{i}$$

7. The symbolic translation $\alpha$ is then obtained as:

$$\alpha_{2tp} = \beta_{2tp} - round(\beta_{2tp}), \alpha_{2tp} \in [-0.5, 0.5)$$

8. The recommended linguistic information is:

$$\left(s_{round(\beta_{2tp})}, \alpha_{2tp}\right)$$

**Table 5** Linguistic Feedback of Two Students in 2-Tuple Form

| Parameter | Feedback of students | | | |
| --- | --- | --- | --- | --- |
| | Student 1 (SS1) | | Student (SS2) | |
| | Linguistic feedback | 2-tuple representation | Linguistic feedback | 2-tuple representation |
| Time taken to solve the question | Small | $(s_1, 0)$ | Large | $(s_3, 0)$ |
| Subject's Knowledge | Large | $(s_3, 0)$ | Limited | $(s_1, 0)$ |
| Liking towards Subject | Moderate | $(s_2, 0)$ | High | $(s_3, 0)$ |
| Perceived preparation level | Moderate | $(s_2, 0)$ | Less | $(s_1, 0)$ |

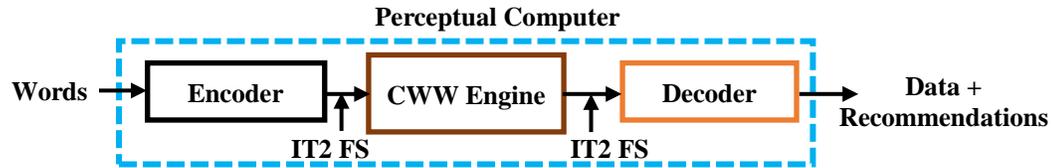

Fig 5: Block diagram of a generic Per-C

## 4 Perceptual computing based CWW framework for student strategy evaluation

The MF of a T1 FS is crisp or precise. It cannot capture the uncertainties associated with the linguistic information in the best possible way because 'words mean different things to different people'. Therefore, the T1 FS based CWW approaches cannot process the linguistic information in the best possible way. Mendel et. al. proposed the concept of IT2 FS based CWW technique called Per-C. Per-C is an instantiation of Zadeh's CWW. At the heart of the Per-C is a block called the Perceptual computer. The block diagram of a generic perceptual computer is shown in Fig. 5. It is mentioned here that a perceptual computer needs to be modified according to the problem in which it is used.

The perceptual computer consists of three building blocks: the Encoder, CWW engine and the Decoder. The task of the encoder is to map the words into their IT2 FS models. The encoder is designed by deciding a problem specific vocabulary of 'words' and then collecting the data intervals from a group of subjects. These data intervals are processed using the Interval Approach (IA) (Liu and Mendel 2008), Enhanced Interval Approach (EIA) (Wu et al. 2012) or the Hao-Mendel Approach (HMA) (Hao and Mendel 2016). All the IA, EIA and HMA (Mendel 2016) have a number of data processing steps divided into two broad categories: the data part and the FS part. The data part consists of numerous steps like bad data processing, outlier processing, etc. FS part consists of calculation of FS models, mapping the IT2 FS models into left, interior or right shoulder FOU, etc. All the processed data values are stored in the form of codebook. The inputs to the encoder are problem specific words and it's output are the IT2 FSs of the words.

Since data comes from diverse sources or different parameters of the same problem, it needs to be aggregated. This task is performed by the CWW engine. It's input are the IT2 FS word models and it's output consists of other IT2 FSs. Other IT2 FSs means that the IT2 FSs at the output of CWW engine are generally different from the ones already stored in the codebook, developed by the encoder. The words may be assigned different weights and these weights may be linguistic.

There are different types of aggregation operators such as interval weighted average (IWA) operator, fuzzy weighted average (FWA) operator and the linguistic weighted average (LWA) operator. The type of aggregation operator used depends on the nature of the data to be aggregated. If at least one of the words or the weights is an IT2 FS, then LWA is used. The aggregation using LWA is given in Eq. (27) as:

$$\tilde{Y}_{LWA} \equiv \frac{\sum_{i=1}^{n} \tilde{X}_i \, \tilde{W}_i}{\sum_{i=1}^{n} \tilde{W}_i} \tag{27}$$

here $\tilde{X}_i$ are the IT2 FS models of the words and $\tilde{W}_i$ are those of weights.

The decoder performs the task of generating the recommendations. There are three types of recommendations that can be generated from the decoder viz. 'word', 'ranking' and the 'class'. 'Ranking' method is used in cases where there are numerous options available at the output and the best alternative amongst them needs to be chosen. Different ranking methods are available of which centroid ranking method is used commonly. The centroid of each $\tilde{Y}_{LWA}$ in Eq. (27) is calculated in the form of an interval $[c_l, c_r]$ using the Enhanced Karnik Mendel (EKM) algorithm as shown in Eq. (28)-(29) as:

$$c_l = \frac{\sum_{i=1}^{L} x_i \bar{\mu}_{\tilde{A}}(x_i) + \sum_{i=L+1}^{N} x_i \underline{\mu}_{\tilde{A}}(x_i)}{\sum_{i=1}^{L} \bar{\mu}_{\tilde{A}}(x_i) + \sum_{i=L+1}^{N} \underline{\mu}_{\tilde{A}}(x_i)} \qquad (28)$$

$$c_r = \frac{\sum_{i=1}^{R} x_i \underline{\mu}_{\tilde{A}}(x_i) + \sum_{i=R+1}^{N} x_i \bar{\mu}_{\tilde{A}}(x_i)}{\sum_{i=1}^{R} \underline{\mu}_{\tilde{A}}(x_i) + \sum_{i=R+1}^{N} \bar{\mu}_{\tilde{A}}(x_i)} \qquad (29)$$

Here, $L$ and $R$ are called switch points. The mean of $c_l$ and $c_r$ is calculated as shown in Eq. (30):

$$c(\tilde{A}) = \frac{c_l + c_r}{2} \qquad (30)$$

'Word' recommendation is generated using Jaccard's similarity methodas shown in Eq. (31):

$$sm_J(\tilde{Y}_{LWA}, \tilde{s}_k)$$
$$= \frac{\sum_{j=1}^{N} \min\left(\bar{\mu}_{\tilde{Y}}(x_j), \bar{\mu}_{\tilde{s}_k}(x_j)\right) + \sum_{j=1}^{N} \min\left(\underline{\mu}_{\tilde{Y}}(x_j), \underline{\mu}_{\tilde{s}_k}(x_j)\right)}{\sum_{j=1}^{N} \max\left(\bar{\mu}_{\tilde{Y}}(x_j), \bar{\mu}_{\tilde{s}_k}(x_j)\right) + \sum_{j=1}^{N} \max\left(\underline{\mu}_{\tilde{Y}}(x_j), \underline{\mu}_{\tilde{s}_k}(x_j)\right)} \qquad (31)$$

here $\tilde{s}_k$ is the IT2 FS word model of the linguistic term. The working of the Per-C is summarized in the form of Algorithm 4.

We now illustrate the student strategy evaluation using the Per-C. First step is to decide the word vocabulary, which is already given in Table 1. Next step is to generate the codebook, which contains the data about the IT2 FS word models. We collected data about the end points of the words shown in Table 1, from a group of subjects. They were asked to provide the end point intervals on scale of 0 to 10 for the words. After processing these data values, the codebook is generated as shown in Table 6. In Table 6, each LMF and UMF is a trapezoid. Also, the LMF and the UMF are the bounding functions for the FOU.

---

**Algorithm 4: Perceptual computing based CWW**

---

1. Build vocabulary of words and ask a group of users to provide the data intervals for the words on scale 0 to 10.
2. Subject the data intervals to IA/ EIA/ HMA. Each of these consists of a data part and a fuzzy set part.
3. Construct word FOUs and store them in the form of codebook.
4. Aggregate data values of the criteria corresponding to user feedback obtained in the gth game in the pth phase (training or execution) at the ith frequency $F_i$ by means of the following Linguistic Weighted Average (LWA):

$$\tilde{Y}_{LWA} \equiv \frac{\sum_{i=1}^{4} \tilde{X}_i \tilde{W}_i}{\sum_{i=1}^{4} \tilde{W}_i}$$

Here, $\tilde{X}_i$ are the IT2 FS models for the user's words and $\tilde{W}_i$ are those of weights assigned to the corresponding words.

5. Calculate the centroid of each $\tilde{Y}_{LWA}$ in the form of interval $[c_l, c_r]$ using the Enhanced Karnik Mendel (EKM) algorithm as:

$$c_l = \frac{\sum_{i=1}^{L} x_i \bar{\mu}_{\tilde{A}}(x_i) + \sum_{i=L+1}^{N} x_i \underline{\mu}_{\tilde{A}}(x_i)}{\sum_{i=1}^{L} \bar{\mu}_{\tilde{A}}(x_i) + \sum_{i=L+1}^{N} \underline{\mu}_{\tilde{A}}(x_i)}$$

$$c_r = \frac{\sum_{i=1}^{R} x_i \underline{\mu}_{\tilde{A}}(x_i) + \sum_{i=R+1}^{N} x_i \bar{\mu}_{\tilde{A}}(x_i)}{\sum_{i=1}^{R} \underline{\mu}_{\tilde{A}}(x_i) + \sum_{i=R+1}^{N} \bar{\mu}_{\tilde{A}}(x_i)}$$

Here, $L$ and $R$ are called switch points.

6. Calculate the mean centroid value from $c_l$ and $c_r$ as:

$$c(\tilde{A}) = \frac{c_l + c_r}{2}$$

7. To generate 'word' recommendation, Jaccard's similarity method is used, as:

$$sm_J(\tilde{Y}_{LWA}, \tilde{s}_k) = \frac{\sum_{j=1}^{N} \min\left(\bar{\mu}_{\tilde{Y}}(x_j), \bar{\mu}_{\tilde{s}_k}(x_j)\right) + \sum_{j=1}^{N} \min\left(\underline{\mu}_{\tilde{Y}}(x_j), \underline{\mu}_{\tilde{s}_k}(x_j)\right)}{\sum_{j=1}^{N} \max\left(\bar{\mu}_{\tilde{Y}}(x_j), \bar{\mu}_{\tilde{s}_k}(x_j)\right) + \sum_{j=1}^{N} \max\left(\underline{\mu}_{\tilde{Y}}(x_j), \underline{\mu}_{\tilde{s}_k}(x_j)\right)}$$

Here $\tilde{Y}_{LWA}$ is given in step 4, and $\tilde{s}_k$ is the IT2 FS word model of the linguistic recommendation, taken form the linguistic term set.

---

**Table 6.** FOU Data values of the Words Shown in Table 1. Each UMF and LMF is a Trapezoid

| Parameter/ Recommendation | Word | LMF | | | | | UMF | | | | Centroid | | Centroid Mean |
|---|---|---|---|---|---|---|---|---|---|---|---|---|---|
| | | e | f | g | i | h | a | b | c | d | $c_l$ | $c_r$ | |
| Time taken to solve the question | Very little (VL) | 0.00 | 0.00 | 0.18 | 2.63 | 0.00 | 0.00 | 0.09 | 1.32 | 1.00 | 0.44 | 0.93 | 0.68 |
| | Small (S) | 0.59 | 2.00 | 3.00 | 4.41 | 1.79 | 2.50 | 2.50 | 3.21 | 0.59 | 1.88 | 3.12 | 2.50 |
| | Moderate (M) | 1.98 | 3.75 | 5.00 | 6.41 | 4.29 | 4.59 | 4.59 | 5.21 | 0.42 | 3.38 | 5.38 | 4.38 |
| | Large (L) | 4.02 | 5.65 | 7.00 | 8.62 | 6.40 | 6.60 | 6.60 | 7.10 | 0.34 | 5.23 | 7.60 | 6.41 |
| | Very Large (VLA) | 6.05 | 9.72 | 10.00 | 10.00 | 8.68 | 9.91 | 10.00 | 10.00 | 1.00 | 8.53 | 9.56 | 9.04 |
| Subject's Knowledge | Very Limited (SVL) | 0.00 | 0.00 | 0.28 | 3.95 | 0.00 | 0.00 | 0.09 | 1.32 | 1.00 | 0.44 | 1.47 | 0.96 |
| | Limited (SL) | 0.59 | 2.00 | 3.00 | 4.41 | 1.79 | 2.50 | 2.5 | 3.21 | 0.59 | 1.88 | 3.12 | 2.50 |
| | Moderate (SM) | 2.38 | 4.5 | 6.50 | 8.62 | 4.9 | 5.32 | 5.32 | 5.6 | 0.26 | 3.62 | 7.29 | 5.46 |
| | Large (SLA) | 4.38 | 6.50 | 8.00 | 9.62 | 6.79 | 7.38 | 7.38 | 8.21 | 0.49 | 6.16 | 8.24 | 7.20 |
| | Very Large (SVLA) | 7.37 | 9.73 | 10.00 | 10.00 | 9.34 | 9.95 | 10.00 | 10.00 | 1.00 | 8.95 | 9.78 | 9.36 |
| Liking towards Subject | Very Less (AVL) | 0.00 | 0.00 | 1.06 | 2.82 | 0.00 | 0.00 | 1.06 | 2.33 | 1.00 | 0.89 | 1.04 | 0.97 |
| | Less (AL) | 2.06 | 2.95 | 3.05 | 5.12 | 2.06 | 2.95 | 3.05 | 4.12 | 1.00 | 3.04 | 3.39 | 3.21 |
| | Moderate (AM) | 3.06 | 4.99 | 5.06 | 7.00 | 3.82 | 4.99 | 5.06 | 6.27 | 1.00 | 4.79 | 5.28 | 5.03 |
| | High (AH) | 5.46 | 6.98 | 7.00 | 8.54 | 5.85 | 6.98 | 7.00 | 8.03 | 1.00 | 6.83 | 7.13 | 6.98 |
| | Very High (AVH) | 7.39 | 8.99 | 10.00 | 10.00 | 7.71 | 8.99 | 10.00 | 10.00 | 1.00 | 9.03 | 9.13 | 9.08 |
| Perceived preparation level | Very Less (PVL) | 0.00 | 0.00 | 1.09 | 2.85 | 0.00 | 0.00 | 1.09 | 2.19 | 1.00 | 0.85 | 1.05 | 0.95 |
| | Less (PL) | 1.21 | 2.99 | 3.03 | 4.94 | 1.69 | 2.99 | 3.03 | 4.24 | 1.00 | 2.82 | 3.21 | 3.02 |
| | Moderate (PM) | 3.50 | 4.99 | 5.03 | 6.85 | 3.8 | 4.99 | 5.03 | 6.24 | 1.00 | 4.92 | 5.22 | 5.07 |
| | High (PH) | 4.97 | 6.98 | 7.03 | 8.28 | 5.89 | 6.98 | 7.03 | 8.19 | 1.00 | 6.72 | 7.06 | 6.89 |
| | Very High (PVH) | 7.03 | 8.98 | 10.00 | 10.00 | 7.62 | 8.98 | 10.00 | 10.00 | 1.00 | 8.92 | 9.10 | 9.01 |
| Strategy of student | Not Good (SSNG) | 0.00 | 0.00 | 1.01 | 2.68 | 0.00 | 0.00 | 1.01 | 2.23 | 1.00 | 0.85 | 0.99 | 0.92 |
| | Below Average (SSBA) | 1.36 | 2.97 | 3.01 | 4.64 | 1.87 | 2.97 | 3.01 | 4.14 | 1.00 | 2.83 | 3.17 | 3.00 |
| | Average (SSA) | 3.42 | 4.95 | 5.01 | 6.37 | 3.97 | 4.95 | 5.01 | 6.18 | 1.00 | 4.86 | 5.1 | 4.98 |
| | Good (SSG) | 4.92 | 6.97 | 7.00 | 9.06 | 5.89 | 6.97 | 7.00 | 8.03 | 1.00 | 6.64 | 7.31 | 6.98 |
| | Very Good (SSVG) | 7.16 | 9.00 | 10.00 | 10.00 | 7.82 | 9.00 | 10.00 | 10.00 | 1.00 | 8.96 | 9.16 | 9.06 |

**Table 7** FOU Data values for Performances of Students. Each UMF and LMF is a Trapezoid

| Student | LMF | | | | | UMF | | | | Centroid | | Centroid Mean | Linguistic recommendation |
|---|---|---|---|---|---|---|---|---|---|---|---|---|---|
| | e | f | g | i | h | a | b | c | d | $c_l$ | $c_r$ | | |
| SS1 | 2.88 | 4.62 | 5.27 | 6.97 | 4.05 | 4.97 | 4.99 | 5.98 | 0.77 | 4.44 | 5.47 | 4.95 | *ssa* or *Average* |
| SS2 | 2.82 | 4.41 | 5.01 | 6.63 | 3.93 | 4.77 | 4.78 | 5.65 | 0.73 | 4.19 | 5.27 | 4.73 | *ssa* or *Average* |

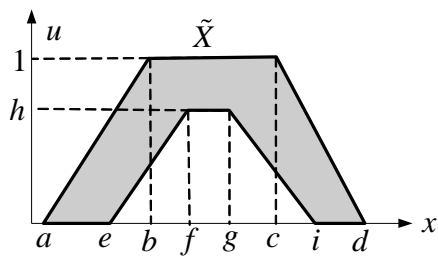

Fig. 6. Parameters of UMF={a, b, c, d} and LMF={ e, f, g, i, h}

Consider the feedback of student SS1 given in Table 2. The FOU data corresponding to the linguistic values of the parameters can be found from Table 6. These are aggregated using the LWA given in Eq. (27). We have assumed equal weights. Therefore, the LWA aggregation is given as:

$$\tilde{Y}_{LWA} = \frac{\tilde{X}_s + \tilde{X}_{sla} + \tilde{X}_{am} + \tilde{X}_{pm}}{4} \qquad (32)$$

The $\tilde{Y}_{LWA}$ in Eq. (32) is an IT2 FS word model. Thus, its centroid values $[c_l, c_r]$, are calculated using the EKM algorithm, using Eqs. (28)-(29) and mean of the centroid, $c(\tilde{A})$, using Eq. (30). The values obtained are:

$$[c_l, c_r] = [4.44, 5.47] \qquad (33)$$

$$c(\tilde{A}) = 4.95 \qquad (34)$$

Using the Jaccard's similarity measure, IT2 FS word model, $\tilde{Y}_{LWA}$, in Eq. (32) corresponding to the strategy of SS1 is compared to the linguistic values of student strategy, given in Table 1. Thus, using Eq. (31), the recommendation generated is: Strategy of SS1= *ssa* or *Average*. Proceeding similarly for SS2, we obtain the data values corresponding to his/ her strategy. All the data values are summarized in Table 7.

It is mentioned here that for the FOU data values given in Tables 6 and 7, each UMF is characterized by four data values whereas each LMF by five, as shown in Fig. 6. The height of UMF is 1. The parameter, '*h*', for the LMF is its height. The last three columns of Table 6 as well as columns 11 to 13 of Table 7, contain the values of two centroid end points and the mean of the centroid, respectively.

## 5 Comparison of recommendations generated by different CWW approaches for Student Strategy evaluation

In our case study, we consider 25 students, in the first semester of Master's course in Computer Science stream at our university. The students study Database Management Systems (DBMS) in the semester, which is a compulsory subject. Students sit its examination at the end of the semester. Students are presented with different questions on varying levels of difficulty, in the examination. They answer the questions to the best of their ability, using their respective strategies.

We have assessed the students' strategies using various CWW approaches viz., extension principle, symbolic method, 2-tuple approach and the Per-C. Each of the CWW approaches generates a strategy score (which is used to compare and rank the respective students based on their strategies) and a linguistic recommendation describing the student strategy.

Table 8 presents the feedback of 25 students for the four parameters as well as the recommendations generated by each CWW approach. Recommendations for each of the CWW approaches have been generated by processing the students' feedbacks in the manner illustrated in Section 3 (for extension principle, symbolic method and 2-tuple approach) and Section 4 (for Per-C).

In Table 8, consider feedback of students 1, 2, 4 to 18, 21, 23 and 25. Even though all these students have a different combination of all four parameters used to assess student's strategy, still the extension principle generates same recommendation of '*average*' student performance (in numeric terms {0.25, 0.5, 0.75}), for all of them. Similarly, consider feedback of students 1, 2, 5, 6, 7, 9 11, 13, 15 and 17. For all these ten students, the symbolic method generates same recommendation of '*average*' student

**Table 8** Comparison of Recommendations Generated by the CWW approaches[a]

| Student | Time taken to solve the question | Subject's Knowledge | Liking towards subject | Perceived preparation level | Extension Principle | | Symbolic Method | | 2-Tuple approach | | Perceptual computing | |
|---|---|---|---|---|---|---|---|---|---|---|---|---|
| | | | | | Numeric | Linguistic | Numeric | Linguistic | Numeric | Linguistic | Numeric | Linguistic |
| 1 | S | SLA | AM | PM | {0.25, 0.5, 0.75} | SSA | 2 | SSA | 2 | SSA | 4.95 | SSA |
| 2 | L | SL | AH | PL | {0.25, 0.5, 0.75} | SSA | 2 | SSA | 2 | SSA | 4.73 | SSA |
| 3 | L | SLA | AVH | PM | {0.5, 0.75, 1} | SSG | 3 | SSG | 3 | SSG | 6.94 | SSG |
| 4 | L | SVLA | AM | PL | {0.25, 0.5, 0.75} | SSA | 3 | SSG | 2.5 | SSG | 5.96 | SSG |
| 5 | S | SVLA | AVL | PM | {0.25, 0.5, 0.75} | SSA | 2 | SSA | 1.75 | SSA | 4.48 | SSA |
| 6 | L | SVLA | AVL | PVL | {0.25, 0.5, 0.75} | SSA | 2 | SSA | 1.75 | SSA | 4.42 | SSA |
| 7 | S | SM | AL | PVH | {0.25, 0.5, 0.75} | SSA | 2 | SSA | 2 | SSA | 5.05 | SSA |
| 8 | VLA | SLA | AH | PL | {0.25, 0.5, 0.75} | SSA | 3 | SSG | 2.75 | SSG | 6.56 | SSG |
| 9 | M | SVLA | AVL | PVL | {0.25, 0.5, 0.75} | SSA | 2 | SSA | 1.5 | SSA | 3.92 | SSA |
| 10 | L | SVL | AVH | PVH | {0.25, 0.5, 0.75} | SSA | 3 | SSG | 2.75 | SSG | 6.37 | SSG |
| 11 | L | SL | AH | PL | {0.25, 0.5, 0.75} | SSA | 2 | SSA | 2 | SSA | 4.73 | SSA |
| 12 | M | SL | AM | PVH | {0.25, 0.5, 0.75} | SSA | 3 | SSG | 2.25 | SSG | 5.23 | SSG |
| 13 | VL | SM | AVH | PM | {0.25, 0.5, 0.75} | SSA | 2 | SSA | 2 | SSA | 5.07 | SSA |
| 14 | L | SVL | AVH | PL | {0.25, 0.5, 0.75} | SSA | 3 | SSG | 2 | SSA | 4.87 | SSA |
| 15 | S | SVL | AL | PVH | {0.25, 0.5, 0.75} | SSA | 2 | SSA | 1.5 | SSA | 3.92 | SSA |
| 16 | VLA | SM | AVL | PVH | {0.25, 0.5, 0.75} | SSA | 3 | SSG | 2.5 | SSG | 6.12 | SSG |
| 17 | VL | SLA | AH | PL | {0.25, 0.5, 0.75} | SSA | 2 | SSA | 1.75 | SSA | 4.47 | SSA |
| 18 | M | SVLA | AM | PM | {0.25, 0.5, 0.75} | SSA | 3 | SSG | 2.5 | SSG | 5.96 | SSG |
| 19 | S | SM | AM | PL | {0, 0.25, 0.5} | SSBA | 1 | SSBA | 1.5 | SSA | 4.00 | SSA |
| 20 | VL | SLA | AL | PL | {0, 0.25, 0.5} | SSBA | 2 | SSA | 1.25 | SSBA | 3.53 | SSBA |
| 21 | S | SL | AVH | PH | {0.25, 0.5, 0.75} | SSA | 3 | SSG | 2.25 | SSA | 5.24 | SSA |
| 22 | VLA | SVL | AVL | PVL | {0, 0.25, 0.5} | SSBA | 1 | SSBA | 1 | SSBA | 2.98 | SSBA |
| 23 | S | SVLA | AH | PVL | {0.25, 0.5, 0.75} | SSA | 3 | SSG | 2 | SSA | 4.97 | SSA |
| 24 | VL | SL | AH | PL | {0, 0.25, 0.5} | SSBA | 2 | SSA | 1.25 | SSBA | 3.30 | SSBA |
| 25 | L | SVL | AVH | PM | {0.25, 0.5, 0.75} | SSA | 3 | SSG | 2.25 | SSA | 5.38 | SSA |

[a] For expanded forms of words in rows 4 to 28 and columns 2 to 5, 7, 9, 11 and 13, refer Table 1

strategy (in numeric terms 2), though all these students have a different combination of all four parameters used to assess student's strategy. Now consider, the case of students 1, 2, 5, 6, 7, 9, 11 to 15, 17, 21, 23 and 25 with 2-tuple approach. For all these students, the same recommendation of '*average*' student performance (in numeric terms 2) is generated, though all these students have a different combination of all four parameters used to assess student's strategy.

On the other hand, none of the cases exists where the perceptual computing gives same recommendation for the students with different combination of all four parameters used to assess student's strategy. For example, for students 1 and 2, the linguistic recommendation is '*average*', but the numeric values are 4.95 and 4.73, respectively.

Thus, from Table 8, it can be seen that in majority cases, the extension principle, symbolic method and 2-tuple approaches fail to generate unique recommendations. Per-C generates unique recommendations in 100% of the cases.

One of the possible reasons for the same is that Per-C captures the 'word' uncertainty in a better manner than the other CWW approaches. The better uncertainty handling capability of Per-C is reflected in the form of IT2 FS word models, which are its objects of computation, thereby resulting in unique recommendations.

Per-C is suitable for all the human driven applications/ environments that involve 'word' modeling for generating recommendations. Other CWW approaches viz., extension principle, symbolic method and 2-tuple approach, will fail to give unique recommendations in such environments.

## 6 Conclusion

CWW is a novel mathematical technique that provides a one-to-one mapping between linguistic and numeric information. Linguistic information is generated in human driven systems and has in-built uncertainty. So the use of CWW to handle such information becomes indispensable. Linguistic information is best modeled by FSs.

The main contribution of this paper is to illustrate the use of CWW for student strategy evaluation and present a comparison of the recommendations generated by different CWW approaches viz., extension principle, symbolic method, 2-tuple approach and perceptual computing. We feel that in any examination, the actual test of a student's ability is not only to solve a question correctly but also with better strategy.

Use of CWW for student strategy evaluation provides us with two major advantages. Firstly, it generates a numeric score corresponding to the strategy adopted by a student in an examination. This enables comparison and ranking of the students based on their performances. Secondly, a linguistic evaluation describing the student strategy is also obtained from the system. Both these numeric score and linguistic recommendation corresponding to the student's strategy, together can be taken as a measure of his/ her learning outcome as perceived by a faculty member. The goal of assessing the learning outcome is to ensure that all the students, especially the ones who have low academic learning, can be paid more attention by the faculty members for improvement.

Furthermore, the linguistic recommendation is useful for human beings as they naturally understand and express themselves using 'words', 'words' being treated as fuzzy information granules in the GC paradigm, which is perhaps the case with most of the human reasoning and concepts.

Our case study involved the comparative analysis of the examination strategy adopted by 25 students, using all the above said CWW approaches. We found that perceptual computing outperformed other CWW techniques (viz., extension principle, symbolic method and 2-tuple) by giving unique recommendations in 100% cases, whereas other techniques failed to do so in majority of the cases.

One of the possible reasons for the same is that perceptual computing captures the 'word' uncertainty in a better manner than the other CWW technique. The better uncertainty handling capability of perceptual computing is reflected in the form of IT2 FS word models, which are its objects of computation, thereby resulting in unique recommendations. The other CWW techniques are not able to distinguish between 'word' uncertainties and depict 'word' models as either uniformly distributed triangular MFs or uniformly spaced linguistic term indices, thus resulting in ambiguous recommendations.

Future work may focus on solving the problem using higher order fuzzy information granules by representing the linguistic information as general type 2 fuzzy sets. Also, the studied CWW approaches may be applied to find solutions in many other real life problems comprising linguistic information.